\pdfoutput=1
\documentclass[11pt, a4paper, twocolumn, copyright]{google}

\usepackage[authoryear, sort&compress, round]{natbib}
\usepackage{booktabs}
\usepackage{multirow}
\usepackage{graphicx}
\usepackage{amsmath}
\usepackage{xcolor}
\usepackage{tikz}
\usepackage{algorithm}
\usepackage[noend]{algpseudocode}
\usetikzlibrary{shapes.geometric, arrows.meta, positioning, calc, fit, backgrounds, decorations.pathreplacing}

\newcommand{\proagent}{Pro\textsuperscript{2} Agent}


\uselogo{} 

\title{\textbf{\proagent}: Scientific Problem Probing for Autonomous AI R$\&$D}

\correspondingauthor{ychen646@jh.edu}

\renewcommand{\today}{2026-11-18}

\author[1,2]{Yixiong Chen\textsuperscript{*}}
\author[1]{Liqian Peng}
\author[1]{Tian Tan}
\author[1]{Ziqian Lin}
\author[1]{Blake Jianhang Chen}
\author[1]{Chao Zhao}
\author[1]{Yaojie Liu}
\author[1]{Alec Go}
\author[2]{Alan Yuille}

\affil[1]{\thepa{}{}}
\affil[2]{Johns Hopkins University}

\begin{abstract}
Large language model (LLM) agents can now autonomously conduct AI research by writing code and running experiments. They typically do so by iteratively optimizing a codebase, while ignoring the problem structure behind the task. This formulation is effective on simpler tasks, but on difficult problems it often gets stuck in the wrong solution region because the agent never explicitly represents its understanding of the problem.
We introduce \proagent{}, which shifts the primary state of autonomous AI research from a codebase state to a Problem State: an explicit structure of what problems prevent success and which solutions are likely to address them. \proagent{} initializes the Problem State through comprehensive analysis of available resources and constraints. It then performs targeted research to relate the identified problems to candidate solutions. Through controlled hypothesis testing, the agent updates the importance of a suspected problem or the benefit brought by a candidate solution.
On the complete MLE-Bench, \proagent{} matches the best reported medal rate (70.7\%) under the standard 24-hour budget and reaches 76.0\% with a 48-hour budget, the highest medal rate reported on the complete benchmark. It also achieves competitive performance on PostTrainBench. In addition, it produces research notes for each task that make the research process transparent and reusable by human researchers.
\end{abstract}

\begin{document}

\maketitle
\begingroup
\renewcommand\thefootnote{*}
\footnotetext{Work done during internship at Google}
\endgroup

\section{Introduction}

Recent advances in large language models (LLMs) have transformed code generation from simple completion into autonomous software engineering. Agents can now operate developer tools, modify repositories, and resolve real-world software issues \citep{swebench,sweagent,openhands}. This progress has motivated growing interest in applying agents to autonomous AI research, including end-to-end developing machine learning (ML) models, LLMs, and their training recipes \citep{mlebench,rebench,paperbench,scicode,posttrainbench}. However, AI research poses challenges beyond software engineering. It relies on expensive and stochastic training with an entire system \citep{mlebench,airadojo,mars,posttrainbench} rather than software whose correctness can be verified by unit tests. An effective AI research agent must navigate the optimization landscape through evaluation and identify directions for improvement behind functional code.

\begin{figure*}[t]
    \centering
    \includegraphics[width=0.8\textwidth]{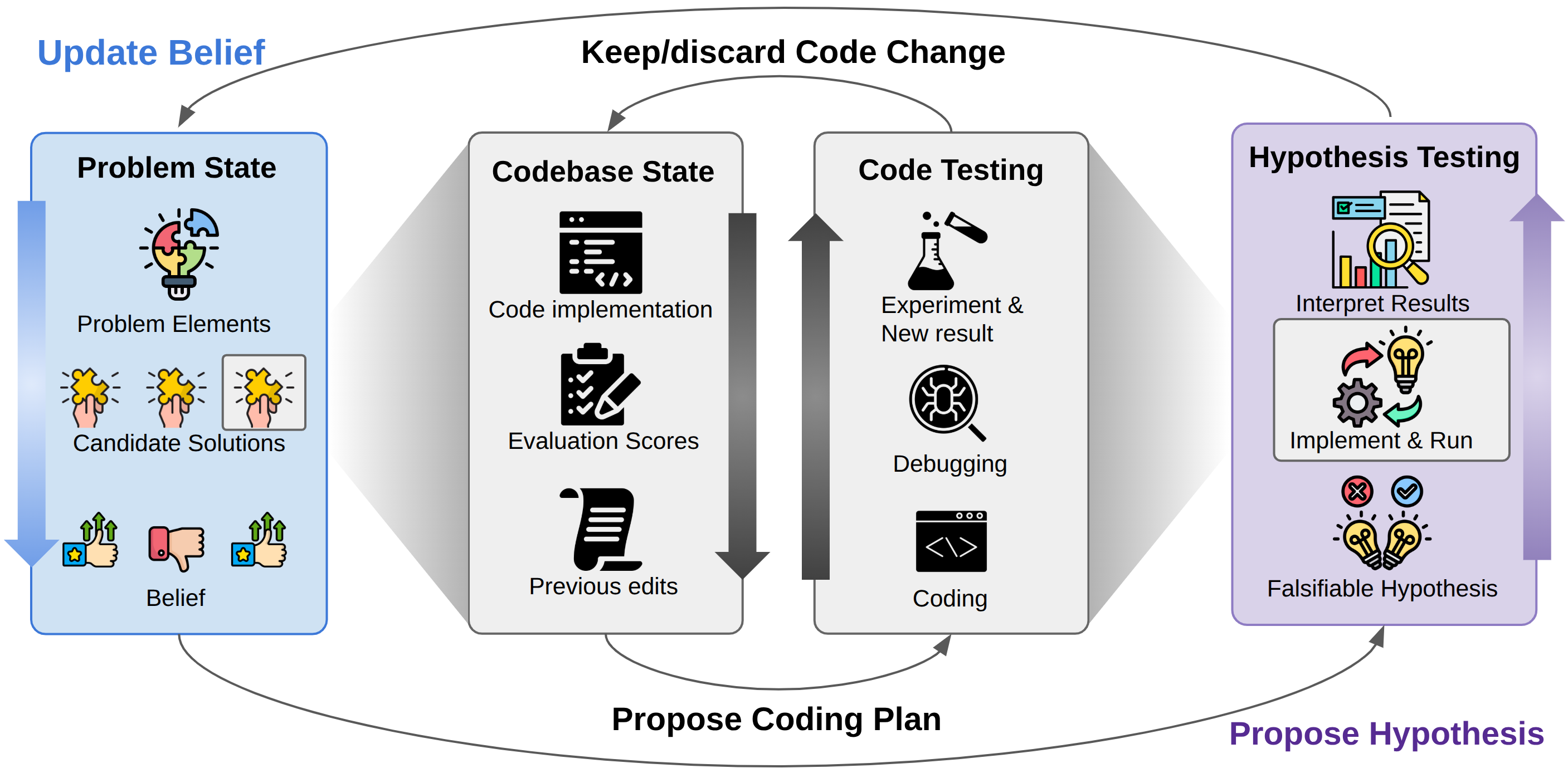}
    \caption{
    \textbf{From codebase state to problem state.}
    Existing AI research agents (gray) mainly iterate between code modification and code testing, using experimental results to select better implementations.
    \proagent{} (full loop, including the blue and purple components) maintains an explicit understanding of the problem and uses experiments to do hypothesis testing to update this understanding and guide future implementations.
    }
    \label{fig:problem_state}
\end{figure*}

Most existing AI research agents address this challenge by formulating model development as search or optimization over candidate code solutions \citep{dsagent,datainterpreter,aide,airadojo,mlestar,mars,rdagent}. AIDE \citep{aide} organizes generated scripts into a solution tree and repeatedly applies drafting, debugging, and improvement operations to promising nodes. AIRA-dojo \citep{airadojo} generalizes this formulation by separating the search policy from the operations that each search can make, enabling controlled comparisons among search/operation strategies. MLE-STAR \citep{mlestar} strengthens the initial solution through web search, then uses ablation studies to identify influential code components for targeted refinement. MARS \citep{mars} extends code search to modular repositories and introduces budget-aware planning and comparative reflective memory for long-running AI research. Other systems use case retrieval, hierarchical agent coordination, or agentic tree search to improve solution generation and execution \citep{dsagent,aibuildai,aiscientistv2}. Despite these differences, their primary research state remains centered on the current codebase, its evaluation score, and the history of previous modifications.

A codebase state is necessary for autonomous AI research, but it provides only a partial view of the research process. It records what has been implemented, how the current solution performs, and which modifications have already been attempted. However, it does not explicitly capture the agent's evolving understanding of why the task remains difficult. As a result, the agent may repeatedly propose locally reasonable code changes while staying within the wrong solution region. In this case, the experimental result is used only to accept or reject a code change, without updating a higher-level explanation of the problem. Figure~\ref{fig:problem_state} illustrates this distinction: codebase-centered agents iterate directly between code modification and code testing, whereas a problem-centered agent runs experiments to provide evidence for hypotheses and update its understanding of the problem before deciding what to implement next.

We introduce \textbf{\proagent{}}, where Pro\textsuperscript{2} stands for \emph{Problem Probing}. Its central idea is to make the agent's understanding of the problem a first-class state of autonomous AI research. We define the \textbf{Problem State} as an explicit representation of what may be limiting performance, which candidate solutions are likely to address it, and their confidence obtained from literature or hypothesis testing. This changes the role of experimentation. Instead of directly searching for the next code modification, \proagent{} uses experiments to test and revise its understanding of the problem. The updated Problem State then guides solution design, allowing the agent to combine evidence-supported treatments for the most important problems into a strong initial solution and to focus later iterations on the remaining uncertainty. In this way, Problem Probing turns a sequence of local code improvements into a directed research process: the agent first determines what needs to be solved, then tests how it should be solved, and only then commits the supported decisions to implementation and optimization.

This formulation provides two practical advantages. First, \proagent{} constructs substantially stronger initial solutions by grounding each major design choice in an explicit understanding of the problem rather than in the language model's immediate preference. Second, it uses hypothesis-driven experiments to reduce uncertainty in the Problem State, directing compute toward a small number of scientifically distinct decisions instead of many local code modifications. On the complete MLE-Bench, \proagent{} achieves a 76.0\% medal rate with a 48-hour budget and 70.7\% with the standard 24-hour budget. On challenging tasks, it reaches medal-level performance within five high-quality scientific iterations, while MARS typically evaluates more than 70 repository variants without consistently reaching comparable scores.\footnote{Counted from the publicly released MARS trajectories in its official repository.} On PostTrainBench, \proagent{}, which uses Gemini 3.5 Flash for implementation and Gemini 3.1 Pro as the supervising PI agent, achieves performance comparable to Claude Code powered by Opus 4.7. These results suggest that stronger autonomous AI research depends not only on deeper solution search or more expensive models, but also on maintaining and continuously improving an explicit understanding of the problem.

Our contributions are threefold:
\begin{enumerate}
    \item We introduce the \textbf{Problem State}, an explicit representation of the agent's understanding of performance-limiting problems, supporting evidence, candidate solutions, and confidence.

    \item We develop \textbf{\proagent{}}, which performs scientific Problem Probing to construct and update the Problem State through evidence and controlled hypothesis testing, then uses it to guide solution design and engineering optimization.

    \item We evaluate \proagent{} on MLE-Bench and PostTrainBench, showing stronger initial solutions, rapid convergence with a small number of scientific iterations, a 76.0\% (48-hour) and 70.7\% (24-hour) medal rate on the complete MLE-Bench, and performance comparable to Claude Code with Opus 4.7 while using Gemini 3.5 Flash as the implementation model.
\end{enumerate}

\section{Related Work}

\paragraph{Automated AI model development.}
Traditional automated machine learning methods reduce human effort by searching over predefined model families, architectures, features, and hyperparameters \citep{autoweka,tpot,autogluon,automlsurvey}. While effective within well-specified search spaces, these methods depend on human expertise to define the candidate space and cannot readily introduce new algorithms or modify complete training systems. Recent large language model agents instead search directly over executable code. AIDE organizes candidate scripts in a solution tree and applies drafting, debugging, and improvement operators to promising nodes \citep{aide}. AIRA-dojo separates the operators that generate solutions from the search policies that select them, enabling systematic comparisons of greedy, evolutionary, and Monte Carlo tree search \citep{airadojo}. AIRA$_2$ further improves experiment throughput, evaluation consistency, and interactive implementation through asynchronous workers and ReAct-based operators \citep{aira2}. Other systems separate research planning from implementation, coordinate specialized agents, or maintain multiple solution trajectories \citep{rdagent,internagent,aibuildai}. These approaches substantially improve autonomous AI model development, but their main search objects remain candidate implementations and their measured performance.

\paragraph{Knowledge-grounded and long-horizon solution search.}
Several AI research agents improve code search using external knowledge, structured memory, or finer search operators. MLE-STAR retrieves task-relevant models from the web to construct its initial solution, then applies ablation-guided refinement to influential code components \citep{mlestar}. AutoMLGen combines a curated domain knowledge base with Monte Carlo Graph Search to reuse information across solution trajectories \citep{automlgen}, while KAPSO organizes external documents, repositories, implementation knowledge, and experimental lessons into a retrievable knowledge system \citep{kapso}. AIBuildAI-2 maintains an evolving hierarchical knowledge base that grounds model design and implementation decisions in externally verifiable sources \citep{aibuildai2}. ML-Master and ML-Master 2.0 focus on accumulating and selecting useful experience over long research horizons \citep{mlmaster,mlmaster2}. MARS searches over modular repositories using budget-aware planning and comparative reflective memory, allowing lessons derived from code and performance differences to guide later solution refinement \citep{mars}. These methods improve the information available to solution search. In contrast, \proagent{} uses external evidence to construct a task-specific model of the problems that must be solved, rather than retrieving knowledge only to propose or refine code.

\paragraph{Hypothesis-driven autonomous AI research.}
A broader line of work aims to automate the scientific process beyond model optimization. The AI Scientist and AI Scientist-v2 generate research ideas, execute experiments, and produce scientific manuscripts, with the latter using agentic tree search to manage experimental branches \citep{aiscientist,aiscientistv2}. CodeScientist combines scientific literature and executable code through genetic search to generate and evaluate research ideas \citep{codescientist}. InternAgent and InternAgent-1.5 organize autonomous discovery around hypothesis generation, verification, evolution, and long-horizon memory \citep{internagent,internagent15}. PiFlow views scientific discovery as principle-guided uncertainty reduction \citep{piflow}, while OR-Agent represents branching research hypotheses and experimental refinements in a structured research tree \citep{oragent}. AutoScientists allows multiple agents to interpret a shared experimental state and organize around promising hypotheses \citep{autoscientists}. SAGE focuses on recovery from failed experiments by maintaining multiple causal explanations and testing them through structured failure attribution \citep{sage}. These systems make hypotheses and experiments more explicit, but generally begin with research ideas or observed experimental failures rather than first constructing a complete representation of what limits performance in the given AI research task.

\paragraph{Explicit research states and evidence-based belief revision.}
Very recent work has begun to externalize scientific reasoning as a persistent state. Arbor introduces Hypothesis Tree Refinement, in which a global research state links hypotheses, executable artifacts, experimental evidence, and reusable insights \citep{arbor}. Its coordinator updates the tree and selects hypotheses, while isolated executors implement and evaluate individual interventions. The Hypothesis Evolution Protocol maintains hypotheses in an auditable registry, associates each hypothesis with evidence and a belief probability, and supports explicit transitions among proposed, supported, refuted, and dormant states \citep{hep}. These methods are closely related to \proagent{} in treating experiments as evidence for updating a higher-level state rather than only as scores for selecting code. However, their state is centered on hypotheses that have already been proposed and the experimental trajectories that test them. \proagent{} instead begins with \emph{Problem Probing}: before solution search, it constructs a Problem State that represents the performance-limiting problems in the task, the evidence for each problem, the candidate solutions associated with it, and the evidence and confidence supporting each solution. The Problem State is first used to synthesize a strong initial solution and is then revised through controlled hypothesis testing. Thus, \proagent{} connects problem diagnosis, solution construction, and iterative experimentation within a single problem-centered representation.

\section{Method}
\label{sec:method}

We present \textbf{\proagent{}}, an autonomous research agent system designed to advance AI research through explicit problem state modeling and hypothesis-driven experimentation. We also integrate a multi-agent design in the system to ensure its implementation capability. We start from formulating the Problem State in Section \ref{subsec:formulation}, and then introduce the three-phase framework in Section \ref{subsec:pro2agent_framework}, finally dive into the details of the multi-agent collaboration in Section \ref{subsec:multi_agent_loop}.

\begin{figure*}[!ht]
    \centering
    \includegraphics[width=\textwidth]{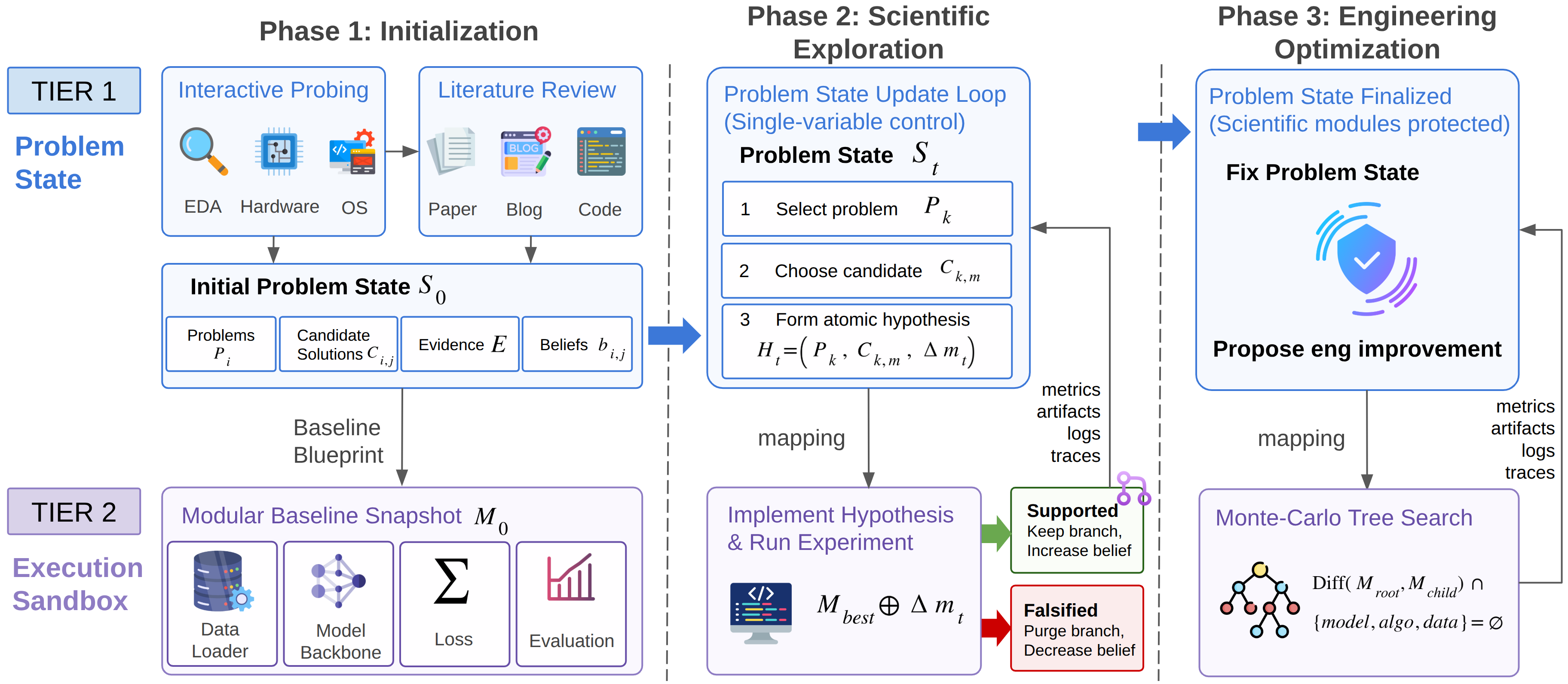}
    \caption{
    \textbf{Overview of the \proagent{} Framework.}
    \proagent{} couples an explicit Problem State with a Git-backed Execution Sandbox across three phases:
    \emph{Initialization} builds the initial problem state and modular baseline;
    \emph{Scientific Exploration} tests single-variable hypotheses and updates evidence-backed beliefs;
    \emph{Engineering Optimization} performs cost-aware MCTS over engineering configurations while preserving the validated architecture.
    }
    \label{fig:framework}
\end{figure*}

\subsection{Problem Formulation}
\label{subsec:formulation}

Formally, traditional AI research agents frame automated model development as code-space optimization: finding an optimal executable script $m^* = \arg\max_{m \in \mathcal{M}} h(m)$, where $\mathcal{M}$ denotes the space of valid code implementations and $h(\cdot)$ is a performance metric evaluated on a validation set. 

In \proagent{}, we redefine the primary state of autonomous AI research from a pure codebase representation to a structured \textbf{Problem State} $\mathcal{S}_t$. At any research step $t$, the Problem State represents the agent's explicit, evolving understanding of the task constraints, performance bottlenecks, candidate solutions, and associated empirical evidence. We express the Problem State as a nested structure:
\begin{equation}
    \mathcal{S}_t = \Big\{ \big(P_i, \mathbf{E}^p_i\big), \big\{ C_{i,j}, b_{i,j}, \mathbf{E}^c_{i,j} \big\}_{j=1}^{L_i} \Big\}_{i=1}^{K}
\end{equation}
where $P_i$ denotes the $i$-th identified performance-limiting problem, and $\mathbf{E}^p_i$ represents its supporting evidence derived from diagnostic profiling. For each problem $P_i$, the agent maintains a list of $L_i$ candidate solutions $C_{i,j}$, each assigned a dynamic belief confidence score $b_{i,j} \in [0, 1]$ and backed by a solution evidence set $\mathbf{E}^c_{i,j}$ compiled from literature or empirical execution traces.

\paragraph{Universal Git-Backed Infrastructure.}
Crucially, every instance of the Problem State $\mathcal{S}_t$ maintains a one-to-one physical mapping with an executable codebase snapshot $M_t \in \mathcal{M}$. To guarantee absolute workspace isolation and prevent state corruption during execution failures, \proagent{} grounds its operational execution on a \emph{Git Version Control System (VCS) infrastructure}. Both the scientific research phase and the engineering optimization phase manage their exploration graph via isolated Git branches, enabling deterministic rollback upon refutation and seamless state recovery.

\subsection{\proagent{} Framework}
\label{subsec:pro2agent_framework}

The full conceptual framework of the agent system is illustrated in Figure \ref{fig:framework}.

\subsubsection{Problem State Initialization}
\label{subsec:initialization}

The initialization phase constructs the initial Problem State $\mathcal{S}_0$ and establishes a robust, modular baseline snapshot $M_0$ through systematic environment probing, literature synthesis, and physical execution validation.

Rather than relying on static metadata, \proagent{} first conducts dynamic runtime profiling via interactive execution probes inside the sandbox. 
By running lightweight diagnostic scripts on high-dimensional raw tensors, the framework extracts actual processing latencies, memory footprints, and domain-specific structural signals, including visual diagnosis for multimodal tasks. 
These empirical measurements are extrapolated mathematically to the full dataset scale against strict compute bounds, proactively unearthing execution bottlenecks $\{P_i\}_{i=1}^K$ such as out-of-memory risks or I/O starvation.

Concurrently, \proagent{} performs a bottleneck-conditioned literature review to seed its initial domain belief bank. 
By mapping identified physical constraints $\{P_i\}_{i=1}^K$ to targeted web and academic search queries, the system retrieves state-of-the-art architectures, domain-specific augmentations, and specialized loss functions. 
Each retrieved solution is structured into candidate hypotheses enriched with literature references and initial belief scores $b_{i,j}$, forming the grounded baseline blueprint and completing the initial Problem State $\mathcal{S}_0$.

Guided by $\mathcal{S}_0$, \proagent{} constructs the modular baseline $M_0$ under strict security and profiling constraints. 
To prevent target leakage and self-grading manipulation, validation evaluation is completely decoupled outside of the experiment sandbox as an HTTP grading service backed by isolated ground-truth labels. 
Furthermore, before committing $M_0$ to the primary Git branch, the framework executes a sanity-check trial on mini-batches to profile runtime health across memory consumption, GPU compute efficiency, and loss stability. 
Once verified under these operational health gates, the baseline $M_0$ is fully executed and committed as the reference anchor for subsequent scientific exploration.

\subsubsection{Scientific Exploration Phase}
\label{subsec:update_loop}

Once $\mathcal{S}_0$ and $M_0$ are established, \proagent{} enters an iterative scientific exploration phase, organized as a resource-bounded best-first search over a tree of hypothesis branches $\mathcal{T}_{sci}$ (Algorithm~\ref{alg:sci_explore}).
At each iteration $t$, the system identifies a target problem $P_k$ from Stage 1 alongside a relevant domain belief to propose a quantitative, atomic hypothesis $\mathcal{H}_t = (P_k, C_{k,m}, \Delta m_t)$, establishing an expected target metric improvement $\delta_{target}$ relative to the current baseline. 
To balance exploration and exploitation across under-investigated bottlenecks, candidate hypotheses in the scientific queue are prioritized using a frequency-penalized score:
\begin{equation}
    \mathrm{Pri}(\mathcal{H}_t) = \frac{b_{k,m}}{N(P_k) + 1}
\end{equation}
where $b_{k,m}$ denotes the associated belief score and $N(P_k)$ counts the historical exploration attempts targeting problem $P_k$. 
This mechanism penalizes repetitive failures on a single bottleneck and forces the framework to broaden its search horizon.

To preserve scientific attribution, \proagent{} strictly enforces single-variable control at both code and architecture levels. 
Upon selecting the highest-priority hypothesis $\mathcal{H}_t$, the system branches an isolated version-controlled environment $\text{\texttt{hypothesis/}}H_t$ from the current best-performing parent node $M_{best}$ in $\mathcal{T}_{sci}$.
The framework then applies the minimal code mutation $\Delta m_t$ and rigorously verifies that the modification strictly adheres to single-variable isolation, rejecting any updates that simultaneously introduce confounding variables unless they are necessary (e.g., batch size or learning rate for another model backbone). 
Once verified, $M_{t+1} = M_{best} \oplus \Delta m_t$ is executed under strict resource watchdogs and a blind grading interface to prevent data leakage.

Following execution, \proagent{} performs a post-hoc qualitative debriefing by evaluating execution telemetry, loss curves, and actual performance gains $\Delta h$. 
The update logic operates on dual fronts depending on the experimental outcome:
\begin{itemize}
    \item \textbf{Supported Hypothesis ($\Delta h \ge \delta_{target}$):} The branch $\text{\texttt{hypothesis/}}H_t$ is retained as a new valid node in $\mathcal{T}_{sci}$ and updates $M_{best}$. The underlying belief score is reinforced via $b_{k,m} \leftarrow \min(1.0, \, b_{k,m} + \Delta b_{pos})$, where $\Delta b_{pos} > 0$.
    \item \textbf{Falsified or Unstable Hypothesis ($\Delta h < \delta_{target}$ or Crash):} The branch $\text{\texttt{hypothesis/}}H_t$ is physically purged via a hard VCS rollback to $M_{best}$. The belief score is penalized via $b_{k,m} \leftarrow \max(0.0, \, b_{k,m} - \Delta b_{neg})$, where $\Delta b_{neg} > 0$.
\end{itemize}
Beyond updating the static belief bank, the framework distills post-hoc insights into a contextual summary that actively propagates through the unexecuted queue. 
By re-evaluating and adjusting the belief scores of remaining pending hypotheses based on newly acquired attribution feedback, \proagent{} dynamically redirects its priority-guided search.

\begin{algorithm}[t]
\caption{Scientific Exploration (Phase 2)}
\label{alg:sci_explore}
\small
\begin{algorithmic}[1]
\Require Problem State $\mathcal{S}_0$, baseline $M_0$, budget $B$
\State $M_{best} \gets M_0$;\; $\mathcal{Q} \gets$ candidate hypotheses from $\mathcal{S}_0$;\; $N(P_k)\gets 0\ \forall k$
\While{budget remains \textbf{and} $\mathcal{Q} \neq \emptyset$}
  \State $\mathcal{H}_t=(P_k, C_{k,m}, \Delta m_t) \gets \arg\max_{\mathcal{H}\in\mathcal{Q}} \mathrm{Pri}(\mathcal{H})$;\; remove $\mathcal{H}_t$ from $\mathcal{Q}$
  \State $N(P_k) \gets N(P_k)+1$
  \State Branch \texttt{hypothesis/}$\mathcal{H}_t$ from $M_{best}$; implement $\Delta m_t$
  \State Reviewer checks single-variable isolation; revise until approved or fall back
  \State Execute $M_{best}\oplus\Delta m_t$ and measure $\Delta h$
  \If{$\Delta h \ge \delta_{target}$}
    \State $M_{best} \gets M_{best}\oplus\Delta m_t$;\; $b_{k,m} \gets \min(1, b_{k,m}+\Delta b_{pos})$
  \Else
    \State Roll back to $M_{best}$;\; $b_{k,m} \gets \max(0, b_{k,m}-\Delta b_{neg})$
  \EndIf
  \State Update $\mathcal{S}_t$ with distilled insights; re-score pending hypotheses in $\mathcal{Q}$
\EndWhile
\State \Return $M_{best}$, $\mathcal{S}_t$
\end{algorithmic}
\end{algorithm}

\subsubsection{Engineering Optimization Phase}
\label{subsec:engineering_optimization}

Upon concluding the scientific exploration phase, \proagent{} transitions to fine-grained engineering optimization to maximize performance from the best validated ML pipeline. 
This phase models hyperparameter tuning, execution optimization, and model ensembling as a classical Monte Carlo Tree Search (MCTS) operating over a symmetric Git-branch tree structure. 
Rooted at the optimal scientific snapshot $M_{best}$, each child node $N_k$ evolves on an isolated branch $\text{\texttt{eng\_}}N_k$, guaranteeing physical reproducibility across competing engineering configurations.

To prevent regression or unverified structural mutations during hyperparameter search, \proagent{} enforces an audit firewall that guarantees strict pipeline invariance:
\begin{equation}
    \mathcal{C}\big(\mathrm{Diff}(M_{root}, M_{child})\big) \cap \{\textsc{model},\textsc{algo},\textsc{data}\} = \emptyset,
\end{equation}
where $\mathcal{C}(\cdot)$ maps a code diff to the set of pipeline components it modifies, as determined by the Reviewer Agent.
The system restricts modifications strictly to engineering variables, such as learning rate schedules, batch sizes, weight decay, and ensemble weights. It rejects any candidate diff that tampers with model architecture, algorithm design, and data representation.

The tree search navigates the engineering space using the Upper Confidence Bound for Trees (UCT) criterion to balance exploration and exploitation. 
Candidate nodes are evaluated using a cost-aware reward function $R(N_k)$ that penalizes computational overhead:
\begin{equation}
    R(N_k) = \Delta h(N_k) - \gamma \ln \big( T_{exec}(N_k) \big)
\end{equation}
where $\Delta h(N_k) = h(M_k) - h(M_{root})$ represents the validation performance gain relative to the root node (with metrics oriented so that higher is better), $T_{exec}(N_k)$ is the execution duration in seconds, and $\gamma > 0$ is a calibrated penalty weight. 
This logarithmic cost penalty suppresses computationally wasteful configurations that yield marginal accuracy gains. 
Upon reaching the allocated engineering budget, the trajectory with the highest validation performance is automatically merged back into the primary Git branch as the final reference model.

\begin{figure*}[t]
    \centering
    \includegraphics[width=\linewidth]{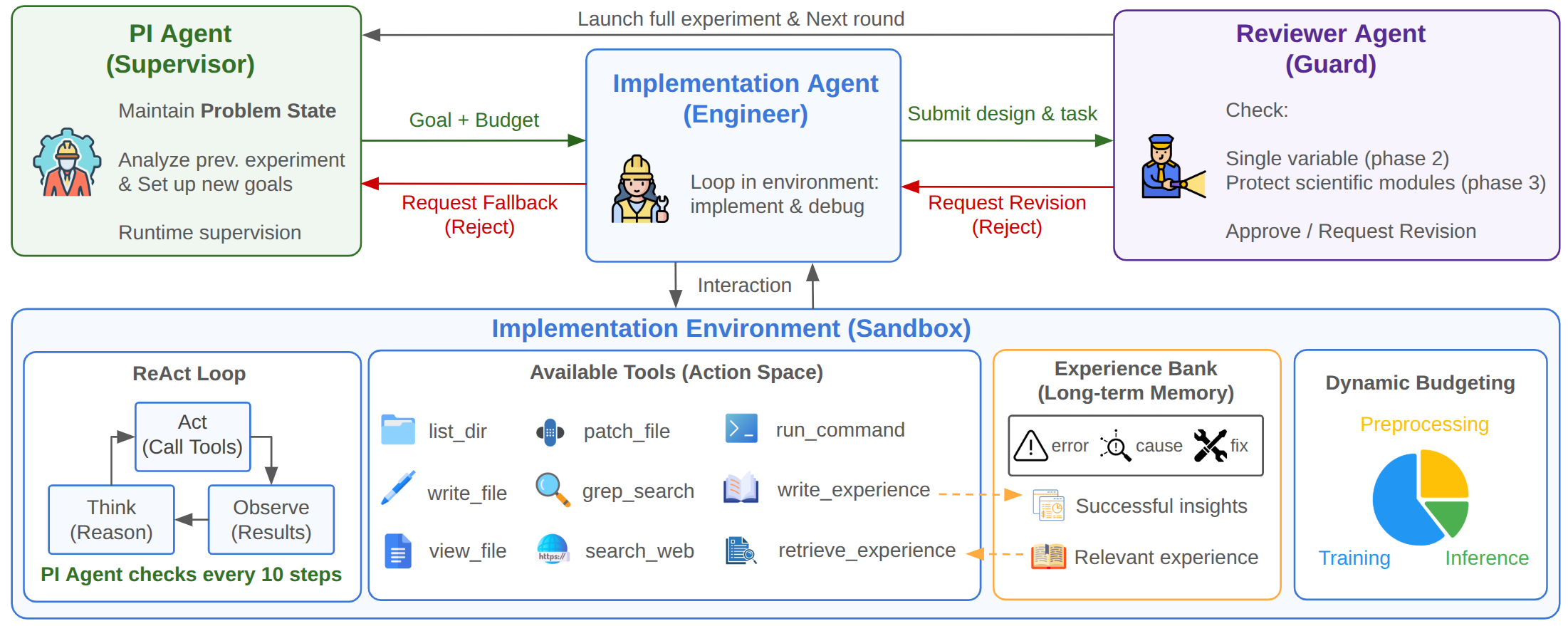}
    \caption{
        \textbf{Tripartite Multi-Agent Loop.} 
        We use a multi-agent engineering loop to ensure every code implementation is as expected without non-compliance and cheating. The PI Agent ($\mathcal{A}_{PI}$) sets goals and acts as a strong-model runtime supervisor, checking execution trajectories every 10 steps to keep the Implementation Agent ($\mathcal{A}_{imp}$) aligned. Once coding and debugging is completed, the Reviewer Agent ($\mathcal{A}_{rev}$) performs code auditing, either approving the submission back to the PI Agent or rejecting it back to the Implementation Agent for targeted remediation.
    }
    \label{fig:react_protocol}
\end{figure*}

\subsection{The Tripartite Multi-Agent Loop}
\label{subsec:multi_agent_loop}

While the high-level framework governs strategic exploration, executing reliable codebase modifications requires strong engineering implementation. We propose a multi-agent loop to ensure engineering needs can be successfully fulfilled without any omission, non-compliance, and cheating. It is invoked for 1) interactive probing (phase 1), 2) baseline construction (phase 1), 3) hypothesis implementation (phase 2), and 4) engineering improvement (phase 3).
As illustrated in Figure~\ref{fig:react_protocol}, \proagent{} establishes a tripartite governance architecture comprising three specialized roles: the \emph{Principal Investigator (PI) Agent} ($\mathcal{A}_{PI}$), the \emph{Implementation Agent} ($\mathcal{A}_{imp}$), and the \emph{Reviewer Agent} ($\mathcal{A}_{rev}$). 
The PI Agent acts as a supervisor that maintains the Problem State, formulates experimental goals alongside time budgets, and provides continuous runtime supervision by inspecting the execution trajectory every 10 steps. 
The Implementation Agent functions as an engineer inside a sandboxed environment, iterating through a ReAct loop (Reasoning, Acting, and Observing) equipped with file and diagnostic tools to implement code mutations and resolve errors. 
To enforce scientific rigor, all modified code must pass through the Reviewer Agent before launching full experiments. 
In Phase 2, the Reviewer enforces single-variable isolation; in Phase 3, it guards against unauthorized mutations to core scientific modules. 
If an implementation fails audit or encounters unresolvable blockers, structured rejection feedback ($\text{\texttt{Request Revision}}$ or $\text{\texttt{Request Fallback}}$) is routed back to trigger targeted remediation or trajectory adjustments.

To ensure long-horizon execution stability and prevent resource exhaustion, \proagent{} integrates a dynamic budgeting protocol enforced alongside an automated monitoring sentinel. 
Computational allocations are dynamically partitioned across three primary execution phases: data preprocessing, model training, and blind evaluation. 
During runtime, the monitoring sentinel polls process telemetry to apply early-stopping policies: halting execution immediately upon loss divergence ($\text{NaN}/\infty$), terminating stalled I/O data loading that exceeds preprocessing caps, or granting soft-margin extensions to late-stage runs near completion to preserve sunk compute investment. 
This dynamic resource governor prevents runaway processes from triggering cluster evictions while maximizing effective compute utilization.

Complementing runtime supervision, the framework incorporates an \emph{Experience Bank} ($\mathcal{E}_{bank}$) within the sandbox to enable persistent long-term memory and cross-iteration self-healing. 
The Experience Bank captures structured engineering lessons structured as error signatures, root causes, and verified fixes. 
Before embarking on a code mutation, $\mathcal{A}_{imp}$ queries relevant historical experiences ($\text{\texttt{retrieve\_experience}}$) to preemptively avoid recurring software bugs or library incompatibilities. 
Upon successfully resolving a novel execution fault, $\mathcal{A}_{imp}$ distills and writes back the debugging insight ($\text{\texttt{write\_experience}}$) to $\mathcal{E}_{bank}$. 
These accumulated reflections are inherited across search phases, allowing the multi-agent framework to prune historically failed parameter regimes and continuously refine its execution velocity.

\section{Experiments}
\label{sec:experiment}

\subsection{Experimental Setup}
\label{sec:experimental_setup}

\paragraph{Datasets.}
We evaluate \proagent{} on two complementary benchmarks. 
MLE-Bench contains 75 Kaggle competitions spanning vision, language, tabular data, time series, signal processing, and structured prediction \citep{mlebench}. 
We evaluate all 75 tasks using one NVIDIA A100 GPU per task under both a 24-hour and a 48-hour wall-clock budget, with one run per budget. 
For controlled studies, we use the official 22-task MLE-Bench Lite split, with a 24-hour budget and three independent runs per task. 
PostTrainBench contains 28 configurations formed by pairing four base language models with seven target benchmarks covering mathematics, science, code generation, function calling, creative writing, and medical dialogue \citep{posttrainbench}. 
Each configuration is run three times, each for 24 hours on one A100 GPU. 
All budgets include analysis, web research, implementation, training, evaluation, and final artifact generation. Unless otherwise noted, the PI Agent uses Gemini 3.1 Pro, and the Implementation and Reviewer Agents use Gemini 3.5 Flash.

\begin{table*}[t]
    \centering
    \caption{
    \textbf{Performance on the complete 75-task MLE-Bench.}
    Published results are from the corresponding papers or the official MLE-Bench leaderboard under their native settings; ``--'' denotes not reported. \proagent{} results are from one run per budget. Following the official leaderboard, submissions that used test-set feedback are not included.
    $^{\ddagger}$Implementation and Reviewer Agents; the PI Agent uses Gemini 3.1 Pro.
    }
    \label{tab:mlebench_full}
    \scriptsize
    \resizebox{\textwidth}{!}{
    \begin{tabular}{llccccccc}
        \toprule
        \textbf{Agent}
        & \textbf{Backbone LLM}
        & \textbf{Compute}
        & \textbf{Valid}
        & \textbf{Above Median}
        & \textbf{Bronze}
        & \textbf{Silver}
        & \textbf{Gold}
        & \textbf{Any Medal} \\
        \midrule
        ML-Master \citep{mlmaster}
        & DeepSeek-R1
        & A100, 12h
        & 93.3
        & 44.9
        & 4.4
        & 7.6
        & 17.3
        & 29.3 \\

        R\&D-Agent \citep{rdagent}
        & GPT-5
        & V100, 12h
        & 53.3
        & 40.4
        & 6.7
        & 12.0
        & 16.4
        & 35.1 \\

        InternAgent \citep{internagent}
        & DeepSeek-R1
        & A800, 12h
        & 96.4
        & 48.4
        & 7.1
        & 10.7
        & 18.7
        & 36.4 \\

        Famou-Agent \citep{fmagent}
        & Gemini 2.5 Pro
        & A800, 24h
        & 96.9
        & 51.6
        & 8.4
        & 12.4
        & 22.7
        & 43.6 \\

        Kapso \citep{kapso}
        & Gemini 3 Pro Preview
        & --, 24h
        & --
        & --
        & --
        & --
        & --
        & 50.7 \\

        ML-Master 2.0 \citep{mlmaster2}
        & DeepSeek-V3.2-Speciale
        & $2\times$4090, 24h
        & 95.6
        & 63.1
        & 11.1
        & 25.8
        & 19.6
        & 56.4 \\

        MARS \citep{mars}
        & Gemini 3 Pro Preview
        & A100, 24h
        & 98.7
        & 65.8
        & 9.3
        & 15.6
        & 31.1
        & 56.0 \\

        MARS+ \citep{mars}
        & Gemini 3 Pro Preview
        & $2\times$H100, 24h
        & 100.0
        & 74.2
        & 12.4
        & 16.4
        & 33.8
        & 62.7 \\

        AIBuildAI \citep{aibuildai}
        & Claude Opus 4.6
        & A100, 24h
        & 100.0
        & 71.1
        & --
        & --
        & 25.8
        & 63.1 \\

        Famou-Agent 2.0 \citep{fmagent}
        & Gemini 3 Pro Preview
        & A800, 24h
        & --
        & --
        & --
        & --
        & --
        & 64.4 \\

        Iris \citep{iris}
        & Claude Opus 4.6
        & --, 12h
        & --
        & --
        & --
        & --
        & --
        & 64.9 \\

        MLEvolve \citep{mlevolve}
        & Gemini 3.1 Pro Preview
        & H200, 12h
        & 100.0
        & 76.0
        & --
        & --
        & 34.7
        & 65.3 \\

        ScienceFlow \citep{scienceflow}
        & DeepSeek-V4-Flash-Preview
        & $\le$2 GPUs, 24h
        & --
        & --
        & --
        & --
        & --
        & 70.2 \\

        AIBuildAI-2 \citep{aibuildai2}
        & Claude Opus 4.7
        & A100, 24h
        & --
        & --
        & --
        & --
        & --
        & 70.7 \\
        \midrule
        
        \textbf{\proagent{}}
        & \textbf{Gemini 3.5 Flash}$^{\ddagger}$
        & \textbf{A100, 24h}
        & 100.0
        & 84.0
        & 25.3
        & 21.3
        & 24.0
        & 70.7 \\

        \textbf{\proagent{}}
        & \textbf{Gemini 3.5 Flash}$^{\ddagger}$
        & \textbf{A100, 48h}
        & 100.0
        & 84.0
        & 29.3
        & 21.3
        & 25.3
        & 76.0 \\
        \bottomrule
    \end{tabular}
    }
\end{table*}

\paragraph{Compared Methods.}
For end-to-end evaluation on the complete MLE-Bench, we compare against published systems that report results on all 75 tasks, including ML-Master, R\&D-Agent, InternAgent, Famou-Agent, Kapso, ML-Master 2.0, MARS, MARS+, AIBuildAI, Famou-Agent 2.0, Iris, MLEvolve, ScienceFlow, and AIBuildAI-2. 
Because these results use different language models, hardware, runtime, and parallelism, they provide system-level context rather than a controlled comparison. 
Our controlled experiments use the same models, tools, implementation agents, software environments, data splits, and training budgets for all matched variants. For MARS, we use the trajectories publicly released in its official repository, which reflect MARS under its native configuration with Gemini 3 Pro Preview. 
We compare \proagent{} with two matched variants: a \emph{Direct Coding Loop} that selects code changes from execution history without a Problem State and a \emph{Search-Augmented Agent} that retrieves relevant methods before doing tasks but does not organize them around diagnosed problems.

\paragraph{Evaluation Metrics.}
On MLE-Bench, we report Valid Submission, Above Median, Bronze, Silver, Gold, and Any Medal rates, with Any Medal Rate as the main metric. On PostTrainBench, we follow the official evaluation and report the weighted average and per-benchmark scores. For controlled studies, we evaluate both the first complete solution and the subsequent search trajectory. Initial quality is measured by medal rates and the normalized bronze gap
\begin{equation}
    G_{\mathrm{bronze}}(s_t)
    =
    \frac{b_t-s_t}{|b_t-m_t|+\epsilon},
\end{equation}
where all task metrics are oriented so that higher is better, and $b_t$ and $m_t$ denote the bronze threshold and leaderboard median. Thus, $G_{\mathrm{bronze}}<0$ indicates medal-level performance. Iteration efficiency is measured by Medal@3, Medal@5 (medal rates at the 3rd and 5th node), and Node AUC. Let $s_{t,k}^{*}$ be the best score found by task $t$ within the first $k$ valid evaluated nodes. We define
\begin{equation}
    \mathrm{NodeAUC}_t
    =
    \frac{1}{T}\sum_{k=1}^{T}
    \frac{s_{t,k}^{*}-b_t}{|b_t-m_t|+\epsilon},
\end{equation}
which rewards methods that reach and maintain strong solutions early along their own search trajectories.

\begin{table*}[t]
    \centering
    \caption{
    \textbf{Performance on the complete PostTrainBench.}
    Published rows are from the official PostTrainBench results.
    They use a 10-hour H100 budget, while \proagent{} uses a 24-hour A100 budget. $^{\ddagger}$Implementation and Reviewer Agents; the PI Agent uses Gemini 3.1 Pro.
    }
    \label{tab:posttrain_full}
    \scriptsize
    \resizebox{\textwidth}{!}{
    \begin{tabular}{lcccccccc}
        \toprule
        \textbf{Method}
        & \textbf{Avg.}
        & \textbf{AIME 2025}
        & \textbf{ArenaHard}
        & \textbf{BFCL}
        & \textbf{GPQA}
        & \textbf{GSM8K}
        & \textbf{HealthBench}
        & \textbf{HumanEval} \\
        
        \midrule

        Official Instruct Models
        & 51.14
        & 29.17
        & 70.21
        & 85.00
        & 36.21
        & 87.00
        & 43.32
        & 71.49 \\
        \midrule
        
        Fable 5 (Claude Code)
        & $41.79 \pm 1.7$
        & $13.33 \pm 2.4$
        & $61.46 \pm 10.8$
        & $73.00 \pm 32.9$
        & $28.35 \pm 2.7$
        & $83.53 \pm 1.9$
        & $40.12 \pm 9.4$
        & $58.38 \pm 13.8$ \\

        GPT 5.6 (Sol, Codex)
        & $36.23 \pm 0.2$
        & $7.50 \pm 3.5$
        & $26.40 \pm 13.5$
        & $94.38 \pm 1.2$
        & $30.13 \pm 1.2$
        & $69.48 \pm 2.7$
        & $27.39 \pm 3.4$
        & $63.95 \pm 2.7$ \\

        Opus 5 (Claude Code)
        & 34.06
        & 8.33
        & 49.58
        & 1.50
        & 30.13
        & 82.96
        & 39.94
        & 53.66 \\

        Opus 4.8 (Claude Code)
        & $32.90 \pm 5.8$
        & $10.83 \pm 2.4$
        & $36.09 \pm 23.0$
        & $47.13 \pm 32.7$
        & $28.35 \pm 2.7$
        & $68.87 \pm 12.9$
        & $31.68 \pm 2.4$
        & $47.33 \pm 22.8$ \\

        Kimi K3 (Claude Code)
        & $31.96 \pm 0.3$
        & $11.66 \pm 1.9$
        & $36.71 \pm 18.9$
        & $9.67 \pm 14.1$
        & $25.23 \pm 4.4$
        & $76.32 \pm 3.1$
        & $31.84 \pm 6.2$
        & $62.30 \pm 3.2$ \\

        GLM 5.2 (Claude Code)
        & $31.70 \pm 2.1$
        & $7.78 \pm 2.1$
        & $26.98 \pm 9.4$
        & $47.92 \pm 39.9$
        & $29.65 \pm 2.0$
        & $72.71 \pm 6.8$
        & $26.55 \pm 5.4$
        & $52.49 \pm 11.8$ \\

        Opus 4.7 (Claude Code)
        & $28.56 \pm 1.7$
        & $6.39 \pm 3.2$
        & $19.59 \pm 18.1$
        & $76.91 \pm 28.0$
        & $27.48 \pm 4.0$
        & $60.72 \pm 17.0$
        & $17.81 \pm 8.7$
        & $42.07 \pm 22.9$ \\

        GPT 5.5 (Codex)
        & $27.23 \pm 0.6$
        & $3.75 \pm 4.1$
        & $12.89 \pm 13.2$
        & $56.13 \pm 49.0$
        & $30.66 \pm 2.2$
        & $56.70 \pm 13.4$
        & $19.79 \pm 3.9$
        & $48.86 \pm 5.9$ \\

        Gemini 3.1 Pro (OpenCode)
        & $21.99 \pm 1.8$
        & $3.89 \pm 1.9$
        & $7.42 \pm 5.4$
        & $63.59 \pm 27.7$
        & $18.53 \pm 8.3$
        & $45.51 \pm 22.3$
        & $16.33 \pm 9.8$
        & $40.19 \pm 8.4$ \\

        \midrule
        \textbf{\proagent{} (Gemini 3.5 Flash)}$^{\ddagger}$
        & \textbf{$30.69 \pm 3.1$}
        & \textbf{$5.75 \pm 4.4$}
        & \textbf{$28.59 \pm 12.2$}
        & \textbf{$78.60 \pm 17.8$}
        & \textbf{$26.45 \pm 3.3$}
        & \textbf{$64.47 \pm 5.1$}
        & \textbf{$19.22 \pm 6.9$}
        & \textbf{$50.63 \pm 16.6$} \\
        \bottomrule
    \end{tabular}
    }
\end{table*}

\subsection{Performance on Complete AI Research Benchmarks}
\label{sec:overall_results}

\paragraph{MLE-Bench.}
Table~\ref{tab:mlebench_full} compares \proagent{} with published systems on all 75 MLE-Bench tasks. \proagent{} produces valid submissions for every task and exceeds the human leaderboard median on 84.0\% of them. It earns 57 medals: 22 bronze, 16 silver, and 19 gold, corresponding to an Any Medal Rate of 76.0\%. Under the standard 24-hour budget, \proagent{} already reaches 70.7\%, matching the strongest previously reported result, AIBuildAI-2 \citep{aibuildai2}, which is powered by Claude Opus 4.7. Extending the budget to 48 hours raises the medal rate to 76.0\%, 5.3 points above AIBuildAI-2 and 13.3 points above MARS+ \citep{mars}, showing that Problem Probing continues to turn additional research time into new medals rather than saturating. Both results are obtained on a single A100 GPU, with the lightweight Gemini 3.5 Flash performing all implementation and review, and Gemini 3.1 Pro acting as the PI Agent.

The improvement is primarily in the breadth of competitive solutions rather than the number of top-ranked solutions. MARS+ achieves a higher Gold Medal Rate than \proagent{} (33.8\% versus 25.3\%), but its combined Bronze and Silver rate is only 28.8\%, compared with 50.7\% for \proagent{}. Similarly, \proagent{} raises the Above Median Rate to 84.0\%, compared with 74.2\% for MARS+ and 76.0\% for MLEvolve. Thus, the main gain comes from converting tasks that previous agents fail to solve competitively into bronze- or silver-level solutions. This result is consistent with the goal of Problem Probing: improving robustness across heterogeneous and difficult tasks, rather than concentrating the search budget on tasks where a strong solution is already easy to identify.

\paragraph{PostTrainBench.}
Table~\ref{tab:posttrain_full} evaluates whether the same research framework transfers from constructing prediction pipelines to post-training language models. \proagent{} obtains a weighted average of $30.69\pm3.1$ across all 28 model--benchmark configurations. Although it remains below the strongest agent result, Fable 5 at $41.79\pm1.7$, it exceeds several systems powered by stronger frontier models, including Opus 4.7 at $28.56\pm1.7$, GPT 5.5 at $27.23\pm0.6$, and Gemini 3.1 Pro at $21.99\pm1.8$. Relative to Opus 4.7 on Claude Code, \proagent{} performs better on five of the seven target benchmarks, with the largest gains on ArenaHard ($28.59$ versus $19.59$) and HumanEval ($50.63$ versus $42.07$), while remaining slightly weaker on AIME 2025 and GPQA. Its gains are therefore distributed across writing, function calling, mathematics, medical dialogue, and code generation rather than arising from a single favorable benchmark.

The two benchmarks require substantially different research outputs: MLE-Bench requires complete data-processing, training, inference, and submission pipelines, whereas PostTrainBench requires the agent to select training data, design a post-training method, and produce a valid model checkpoint. Strong performance in both settings indicates that \proagent{} is not limited to competition-specific code optimization. At the same time, the gap from the official instruction-tuned models on PostTrainBench ($30.69$ versus $51.14$) shows that broad autonomous post-training remains unsolved. Since published systems use different models and compute settings, these tables establish end-to-end capability and cross-domain generality; the following controlled experiments isolate whether these gains arise from Problem Probing and hypothesis-guided iteration.

\subsection{Problem Probing Produces Stronger Starts and Faster Convergence}
\label{sec:initialization_and_iteration}

We next examine how Problem Probing changes the entire research trajectory, from the first complete solution to subsequent score-guided improvement. Unlike a comparison from a shared root, each method constructs its own initial solution and follows its own sequence of evaluated nodes. We therefore evaluate both the quality of the first valid solution and the performance reached after a fixed number of valid evaluated nodes.

\begin{figure*}[t]
    \centering
    \includegraphics[width=\textwidth]{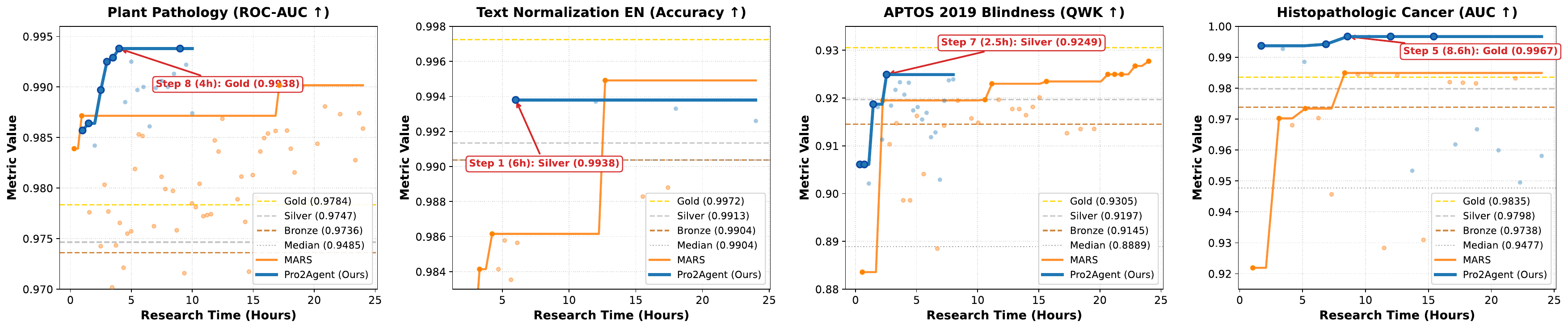}
    \caption{
    \textbf{\proagent{} starts higher and reaches medal thresholds earlier.}
    Validation trajectories on four diverse MLE-Bench tasks plotted against wall-clock search time (hours).
    The solid blue lines represent \proagent{}'s optimization trajectory, while the orange lines represent MARS. MARS trajectories are taken from its public repository. Horizontal dashed lines indicate the corresponding Kaggle leaderboard medal thresholds.
    On all four tasks, \proagent{} starts from a stronger first solution and reaches its medal tier earlier than MARS. Higher is better for all four metrics.
    }
    \label{fig:trajectory_comparison}
\end{figure*}

\begin{table*}[t]
    \centering
    \caption{
    \textbf{Initial solution quality and end-to-end node efficiency on the 22-task MLE-Bench Lite split.}
    The left block evaluates the first complete solution before score-guided revision.
    The right block reports the best Any Medal Rate reached within the first three and five valid evaluated nodes along each method's own trajectory.
    Node indices are method-specific, and the methods do not begin from a shared root. MARS results are computed from its publicly released trajectories.
    Rates are percentages; results are reported as mean $\pm$ standard error over three runs.
    }
    \label{tab:initialization_and_efficiency}
    \scriptsize
    \resizebox{\textwidth}{!}{
    \begin{tabular}{lcccccccc}
        \toprule
        & \multicolumn{5}{c}{\textbf{First Complete Solution (Baseline)}}
        & \multicolumn{3}{c}{\textbf{Own Search Trajectory}} \\
        \cmidrule(lr){2-6}
        \cmidrule(lr){7-9}
        \textbf{Method}
        & \textbf{Above Median}
        & \textbf{Any Medal}
        & \textbf{Gold}
        & $\boldsymbol{G_{\mathrm{bronze}}} \downarrow$
        & \textbf{Build Time (min)}
        & \textbf{Medal@3}
        & \textbf{Medal@5}
        & \textbf{Node AUC} $\uparrow$ \\
        \midrule

        Direct Coding Loop
        & $25.8 \pm 1.5$
        & $12.1 \pm 3.0$
        & $4.5 \pm 0.0$
        & $1.56 \pm 0.09$
        & $2.1 \pm 1.2$
        & $21.2 \pm 3.0$
        & $30.3 \pm 3.0$
        & $-0.60 \pm 0.07$ \\

        Search-Augmented
        & $39.4 \pm 3.0$
        & $28.8 \pm 4.0$
        & $12.1 \pm 1.5$
        & $1.23 \pm 0.07$
        & $8.4 \pm 2.3$
        & $30.3 \pm 1.5$
        & $36.4 \pm 5.5$
        & $-0.44 \pm 0.12$ \\

        MARS
        & $50.0 \pm 2.6$
        & $28.8 \pm 4.0$
        & $12.1 \pm 1.5$
        & $0.96 \pm 0.20$
        & $14.8 \pm 2.6$
        & $34.8 \pm 3.0$
        & $39.4 \pm 5.5$
        & $-0.39 \pm 0.09$ \\
        \midrule

        \textbf{\proagent{}}
        & $\mathbf{81.8 \pm 2.6}$
        & $\mathbf{57.6 \pm 1.5}$
        & $\mathbf{27.3 \pm 0.0}$
        & $\mathbf{-0.57 \pm 0.05}$
        & $11.2 \pm 4.4$
        & $\mathbf{63.6 \pm 2.6}$
        & $\mathbf{66.7 \pm 1.5}$
        & $\mathbf{0.58 \pm 0.17}$ \\
        \bottomrule
    \end{tabular}
    }
\end{table*}

Figure~\ref{fig:trajectory_comparison} gives a direct view of this behavior across four diverse and highly specialized domains (e.g., NLP, standard computer vision, and specialized medical imaging) mapped against actual wall-clock search time. Across all cases, \proagent{} starts from a stronger first solution and reaches medal-level performance substantially earlier than MARS, often within a few hours. On \emph{Plant Pathology}, while both methods' initial valid solutions land above the Gold threshold, \proagent{} steadily optimizes to a stable 0.9938 peak. In contrast, MARS's exploration exhibits high variance, frequently proposing degraded solutions that fall below Gold. On \emph{Text Normalization EN}, \proagent{} achieves a Silver-medal-level score (0.9938) with its first hypothesis at 6 hours, whereas MARS needs roughly 12 hours of exploration to cross the medal thresholds. On \emph{APTOS 2019 Blindness}, \proagent{} starts at a high baseline and crosses the Silver threshold in 2.5 hours; although MARS's trajectory visually appears to reach the Silver line simultaneously, it actually plateaus at 0.9194---strictly below the 0.9196 Silver threshold---for several hours, crossing it only after about 11 hours, more than four times later than \proagent{}. Finally, on \emph{Histopathologic Cancer}, although both methods eventually plateau, their early trajectories differ drastically. MARS spends its first 8 hours exploring suboptimal, low-scoring solutions before reaching the Gold plateau, whereas \proagent{}'s very first evaluated hypothesis (at 1.7 hours) lands directly at an exceptional 0.9937, far above the Gold limit. This illustrates how Problem Probing helps the agent avoid spending its early budget in suboptimal regions. Across these cases, Problem Probing provides a stronger starting point, and hypothesis-guided experiments close the remaining gap within a fraction of the time budget.

Table~\ref{tab:initialization_and_efficiency} confirms that the trajectory patterns extend across the full MLE-Bench Lite split. Before any score-guided revision, \proagent{} already reaches an Any Medal Rate of 57.6\%, twice that of Search-Augmented and MARS initialization at 28.8\%. Its first solutions also exceed the leaderboard median on 81.8\% of tasks and obtain gold medals on 27.3\%, compared with 50.0\% and 12.1\% for MARS. The normalized bronze gap further changes from $0.96$ for MARS to $-0.57$ for \proagent{}, meaning that the average first solution moves from below the bronze boundary to inside the medal region. This gain is not explained by longer initialization: \proagent{} requires less construction time than MARS on average.

The advantage persists as each method follows its own search trajectory. Within three evaluated nodes, \proagent{} reaches a 63.6\% medal rate, compared with 34.8\% for MARS; after five nodes, the rates are 66.7\% and 39.4\%, respectively. The higher Node AUC shows that \proagent{} maintains stronger solutions throughout the search rather than obtaining its advantage only at the final node. These results support a combined account of the improvement: Problem Probing places the first solution in a more suitable region of the solution space, while hypothesis-guided iteration spends the remaining evaluations on unresolved task bottlenecks instead of repairing basic mismatches in data representation, model choice, or evaluation design.

\subsection{Ablation of the Problem State}
\label{sec:ablation}

Table~\ref{tab:problem_state_ablation} separates the contributions of constructing, grounding, and updating the Problem State. Replacing the Problem State with a Codebase State substantially weakens both initialization and final performance. Constructing a Problem State without evidence grounding recovers much of this loss, but remains less reliable because diagnosed bottlenecks and candidate solutions cannot be distinguished from unsupported language-model suggestions. A grounded but static Problem State produces a nearly identical initial solution to the full method, yet cannot incorporate later experimental evidence into subsequent research decisions.

Under the complete 24-hour budget, the full \proagent{} reaches medal-level performance on 20 of the 22 MLE-Bench Lite tasks in two of the three runs and on 19 in the third (89.4\% on average). The Static Problem State averages 83.3\%, about 1.3 fewer medaled tasks per run, showing that the Problem State is not only an initialization artifact. Continually revising the state allows later experiments to update solution beliefs, reject unsupported explanations, and redirect the remaining budget toward unresolved bottlenecks.

\begin{table}[t]
    \centering
    \caption{
    \textbf{Ablation of the Problem State on MLE-Bench Lite.}
    Initial Medal measures the first complete solution.
    Final Medal measures the best solution reached within the full 24-hour budget (mean $\pm$ standard error over 3 runs).
    Full-Trajectory Node AUC summarizes the normalized best-score trajectory over all valid evaluated nodes.
    }
    \label{tab:problem_state_ablation}
    \scriptsize
    \resizebox{\columnwidth}{!}{
    \begin{tabular}{lccc}
        \toprule
        \textbf{Variant}
        & \textbf{Initial Medal}
        & \textbf{Final Medal}
        & \textbf{Node AUC} $\uparrow$ \\
        \midrule

        Codebase State
        & $10.6 \pm 1.5$
        & $56.1 \pm 3.0$
        & $0.06$ \\

        Prob. State w/o Grounding
        & $47.0 \pm 1.5$
        & $77.3 \pm 2.6$
        & $0.27$ \\

        Static Prob. State
        & $56.1 \pm 1.5$
        & $83.3 \pm 3.0$
        & $0.45$ \\

        \textbf{Full \proagent{}}
        & $\mathbf{57.6 \pm 1.5}$
        & $\mathbf{89.4 \pm 1.5}$
        & $\mathbf{0.66}$ \\
        \bottomrule
    \end{tabular}
    }
\end{table}

\begin{figure*}[t]
    \centering
    \includegraphics[width=\textwidth]{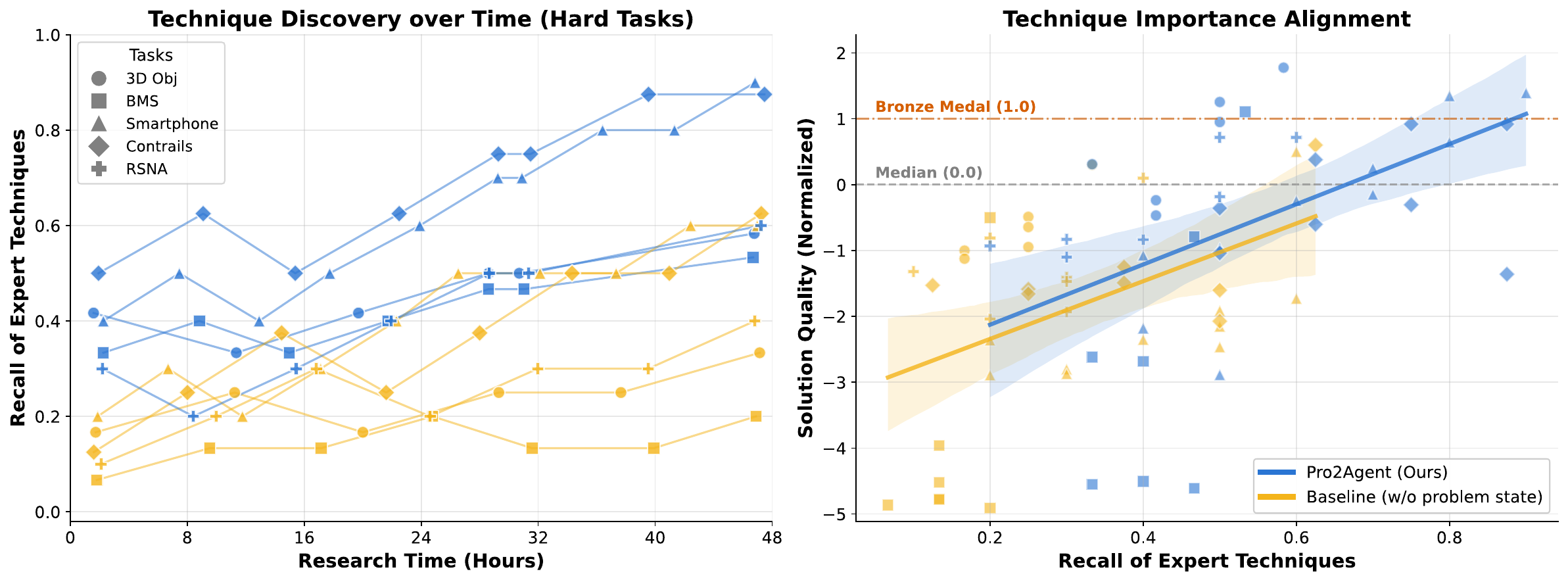}
    \caption{
        \textbf{Expert Technique Discovery on Hard Tasks.} 
        \textit{Left:} Technique recall over a 48-hour search budget. \proagent{} (blue) recalls more expert techniques over time than the baseline without a Problem State (yellow); recall can drop when a falsified hypothesis is rolled back.
        \textit{Right:} Correlation between technique recall and normalized solution quality. At comparable recall, \proagent{} tends to reach higher solution quality, suggesting that the techniques it adopts are better aligned with the diagnosed bottlenecks.
    }
    \label{fig:technique_recall}
\end{figure*}

\subsection{Analysis of Expert Technique Discovery}
\label{sec:technique_discovery}

To further understand how the Problem State aids in searching the hypothesis space on extremely difficult tasks, we analyze the discovery of expert-level techniques (Figure~\ref{fig:technique_recall}). We track the recall of known expert techniques (derived from Kaggle winning solutions) and correlate them with the normalized quality of the solutions over time across the 5 hardest tasks. 

As shown in the left subplot of Figure~\ref{fig:technique_recall}, our method (\proagent) discovers a higher fraction of expert techniques on average compared to the baseline without a Problem State. \proagent{} starts with higher technique recall on every task and keeps a higher recall throughout the search, even though recall occasionally drops when a falsified hypothesis is rolled back. 

Furthermore, the right subplot illustrates that the solution quality of \proagent{} is overall higher than the baseline across all recall levels. Both regression lines show a positive correlation, confirming that discovering more expert techniques generally leads to better performance. The regression line for our method lies above the baseline across the observed recall range. This suggests that the Problem State does not merely increase the number of techniques collected, but guides the agent toward techniques that address the diagnosed bottlenecks, yielding higher performance at a similar level of technique recall.

\subsection{Case Studies: Problem Diagnosis, Scientific Iteration, and Remaining Limits}
\label{sec:case_studies}

To examine how \proagent{} conducts autonomous research, we analyze its qualitative trajectories across four representative tasks spanning both MLE-Bench and PostTrainBench. Prior autonomous systems such as AIDE \citep{aide} and MARS \citep{mars} organize long-horizon exploration primarily around candidate code implementations, using reflective memory to compare score changes across script revisions. In contrast, \proagent{} additionally maintains an explicit, task-level \textbf{Problem State} ($P$) that tracks suspected performance bottlenecks, supporting evidence, and hypothesis status. The essential question is therefore how the agent's understanding of the task evolves over time: what bottlenecks were initially diagnosed ($P_0$), how controlled experiments ($H_1, H_2$) altered those beliefs, and how the system navigated both successful resolutions and hard task boundaries.

\begin{figure*}[t]
\centering
\resizebox{\textwidth}{!}{%
\begin{tikzpicture}[
font=\sffamily\scriptsize,
state/.style={draw=black!60,fill=white,rounded corners=3pt,text width=4.0cm,align=left,inner sep=3.5pt,line width=0.6pt},
resolved/.style={draw=teal!80!black,fill=teal!6,rounded corners=3pt,text width=4.0cm,align=left,inner sep=3.5pt,line width=0.8pt},
failed/.style={draw=gray!70!black,fill=gray!8,rounded corners=3pt,text width=4.0cm,align=left,inner sep=3.5pt,line width=0.8pt},
exp/.style={draw=blue!45,fill=blue!4,rounded corners=2pt,text width=2.15cm,align=center,inner sep=2.5pt,line width=0.6pt,font=\sffamily\tiny},
arrow/.style={->,>=Stealth,color=blue!70!black,line width=0.8pt},
title/.style={font=\sffamily\bfseries\footnotesize,align=left}
]

\node[title] (ta) at (0,0)
{\textbf{(a) MLE-Bench Success: BMS Molecular Translation (Input: molecular image; Output: molecular formula)}};

\node[state,below=0.18cm of ta.south west,anchor=north west] (a0) {
\textbf{Initial Problem State ($P_0$)}\\
$\bullet$ \textbf{Bond Detail Loss:} Typical $224\times224$ downsampling erases stereocenters
\textcolor{red!70!black}{[\textbf{High}]}\\
$\bullet$ \textbf{InChI Syntax Fragility:} Autoregressive errors corrupt full sequence
\textcolor{red!70!black}{[\textbf{High}]}
};

\node[exp,right=0.22cm of a0] (ae1) {
\textbf{$H_1$}\\
Crop + 384px\\[1pt]
$\Delta$:
$>15.0 \rightarrow 4.61$
};

\node[state,right=0.22cm of ae1] (a1) {
\textbf{Intermediate State ($P_1$)}\\
$\bullet$ \textbf{Bond Detail Loss:} Mitigated via patch tokenization
\textcolor{teal!70!black}{[\textbf{Low} $\downarrow$]}\\
$\bullet$ \textbf{InChI Syntax Fragility:} Remains dominant
\textcolor{red!70!black}{[\textbf{High} $\uparrow$]}
};

\node[exp,right=0.22cm of a1] (ae2) {
\textbf{$H_2$}\\
Valency Beam Search\\[1pt]
$\Delta$:
$4.61 \rightarrow 1.7257$
};

\node[resolved,right=0.22cm of ae2] (a2) {
\textbf{Final State ($P_2$): \textcolor{teal!80!black}{Resolved}}\\
$\bullet$ \textbf{Constrained Decoding:} Valency rules eliminate syntax errors
\textcolor{teal!70!black}{[\textbf{Resolved}]}\\
\textbf{Outcome:} Bronze Medal\\
\textbf{Score:} 1.7257 (val) / 1.8860 (test)
};

\draw[arrow] (a0.east)--(ae1.west);
\draw[arrow] (ae1.east)--(a1.west);
\draw[arrow] (a1.east)--(ae2.west);
\draw[arrow] (ae2.east)--(a2.west);

\begin{scope}[on background layer]
\node[draw=black!20,fill=gray!2,rounded corners=5pt,inner sep=5pt,
fit=(ta)(a0)(ae1)(a1)(ae2)(a2)] (fa) {};
\end{scope}

\node[title,below=0.55cm of fa.south west,anchor=north west] (tb)
{\textbf{(b) MLE-Bench Failure: RSNA Breast Cancer Detection (Input: mammography; Output: probability of cancer)}};

\node[state,below=0.18cm of tb.south west,anchor=north west] (b0) {
\textbf{Initial Problem State ($P_0$)}\\
$\bullet$ \textbf{Micro-calcification:} Tiny lesion $<0.1\%$ pixels in $3\text{k}\times2\text{k}$ DICOM
\textcolor{red!70!black}{[\textbf{High}]}\\
$\bullet$ \textbf{Class Imbalance:} $<2\%$ positive cases
\textcolor{red!70!black}{[\textbf{High}]}
};

\node[exp,right=0.22cm of b0] (be1) {
\textbf{$H_1$}\\
Convnext \texttt{nano}\\
1024px + Focal Loss\\[1pt]
$\Delta$:
$0.2577 \rightarrow 0.3514$
};

\node[state,right=0.22cm of be1] (b1) {
\textbf{Intermediate State ($P_1$)}\\
$\bullet$ \textbf{Micro target \& Imbalance:} Resolved via 1024px \& Focal loss
\textcolor{teal!70!black}{[\textbf{Resolved} $\downarrow$]}\\
$\bullet$ \textbf{[New] Model Capacity:} Nano model is weak
\textcolor{red!70!black}{[\textbf{High} $\uparrow$]}
};

\node[exp,right=0.22cm of b1] (be2) {
\textbf{$H_2$}\\
Convnext \texttt{tiny}\\[1pt]
$\Delta$: Instability /\\
Runtime $>$ budget
};

\node[failed,right=0.22cm of be2] (b2) {
\textbf{Final State ($P_2$): \textcolor{gray!80!black}{Resource-Limited}}\\
$\bullet$ \textbf{Capacity Scaling:} Plausible but hits 4h training budget limit
\textcolor{gray!80!black}{[\textbf{Bound} $\uparrow$]}\\
\textbf{Outcome:} Above Median\\
\textbf{Score:} 0.3514 (pF1)
};

\draw[arrow] (b0.east)--(be1.west);
\draw[arrow] (be1.east)--(b1.west);
\draw[arrow] (b1.east)--(be2.west);
\draw[arrow] (be2.east)--(b2.west);

\begin{scope}[on background layer]
\node[draw=black!20,fill=gray!2,rounded corners=5pt,inner sep=5pt,
fit=(tb)(b0)(be1)(b1)(be2)(b2)] (fb) {};
\end{scope}

\node[title,below=0.55cm of fb.south west,anchor=north west] (tc)
{\textbf{(c) PostTrainBench Success: Gemma3-4B GSM8K (Grade school math word problems)}};

\node[state,below=0.18cm of tc.south west,anchor=north west] (c0) {
\textbf{Initial Problem State ($P_0$)}\\
$\bullet$ \textbf{Reasoning Ability:} Pretrained model's output lacks CoT
\textcolor{red!70!black}{[\textbf{High}]}\\
$\bullet$ \textbf{Arithmetic Context Drift:} Logic diverges over 2--8 steps
\textcolor{red!70!black}{[\textbf{High}]}
};

\node[exp,right=0.22cm of c0] (ce1) {
\textbf{$H_1$}\\
CoT SFT\\
reasoning traces\\[1pt]
$\Delta$:
$0.0\% \rightarrow 52.50\%$
};

\node[state,right=0.22cm of ce1] (c1) {
\textbf{Intermediate State ($P_1$)}\\
$\bullet$ \textbf{Reasoning Ability:} Structured CoT becomes stable
\textcolor{teal!70!black}{[\textbf{Resolved} $\downarrow$]}\\
$\bullet$ \textbf{Arithmetic Context Drift:} Policy drift across rollouts
\textcolor{red!70!black}{[\textbf{High} $\uparrow$]}
};

\node[exp,right=0.22cm of c1] (ce2) {
\textbf{$H_2$}\\
GRPO with SymPy\\ 
verification\\
$\Delta$:
$52.5\% \rightarrow 56.5\%$
};

\node[resolved,right=0.22cm of ce2] (c2) {
\textbf{Final State ($P_2$): \textcolor{teal!80!black}{Resolved}}\\
$\bullet$ \textbf{Verifiable RL Policy:} GRPO with mathematical equivalence reward
\textcolor{teal!70!black}{[\textbf{Resolved}]}\\
\textbf{Outcome:} Successful Post-Training\\
\textbf{Score:} 56.50\%
};

\draw[arrow] (c0.east)--(ce1.west);
\draw[arrow] (ce1.east)--(c1.west);
\draw[arrow] (c1.east)--(ce2.west);
\draw[arrow] (ce2.east)--(c2.west);

\begin{scope}[on background layer]
\node[draw=black!20,fill=gray!2,rounded corners=5pt,inner sep=5pt,
fit=(tc)(c0)(ce1)(c1)(ce2)(c2)] (fc) {};
\end{scope}

\node[title,below=0.55cm of fc.south west,anchor=north west] (td)
{\textbf{(d) PostTrainBench Failure: Qwen3-1.7B AIME 2025 (American Invitational Mathematics Examination)}};

\node[state,below=0.18cm of td.south west,anchor=north west] (d0) {
\textbf{Initial Problem State ($P_0$)}\\
$\bullet$ \textbf{Format / Extraction Fragility:} Regex parsing failure
\textcolor{red!70!black}{[\textbf{High}]}\\
$\bullet$ \textbf{Deductive Reasoning Breakdown:} Algebraic step failure
\textcolor{red!70!black}{[\textbf{High}]}
};

\node[exp,right=0.22cm of d0] (de1) {
\textbf{$H_1$}\\
CoT SFT +\\
Parser\\[1pt]
Extraction errors\\
$\rightarrow 0\%$
};

\node[state,right=0.22cm of de1] (d1) {
\textbf{Intermediate State ($P_1$)}\\
$\bullet$ \textbf{Format / Extraction:} Eliminated via SFT and parser
\textcolor{teal!70!black}{[\textbf{Resolved} $\downarrow$]}\\
$\bullet$ \textbf{Core Olympiad Reasoning:} Dominant unresolved bottleneck
\textcolor{red!70!black}{[\textbf{High} $\uparrow$]}
};

\node[exp,right=0.22cm of d1] (de2) {
\textbf{$H_2$}\\
GRPO Policy\\
Optimization\\[1pt]
$\Delta$: 0 successful\\
traces sampled
};

\node[failed,right=0.22cm of de2] (d2) {
\textbf{Final State ($P_2$): \textcolor{gray!80!black}{Unresolved}}\\
$\bullet$ \textbf{RL Sampling Failure:} No reward signal found
\textcolor{gray!80!black}{[\textbf{Unresolved}]}\\
\textbf{Outcome:} Unresolved Bottleneck\\
\textbf{Score:} 6.67\% (2/30)
};

\draw[arrow] (d0.east)--(de1.west);
\draw[arrow] (de1.east)--(d1.west);
\draw[arrow] (d1.east)--(de2.west);
\draw[arrow] (de2.east)--(d2.west);

\begin{scope}[on background layer]
\node[draw=black!20,fill=gray!2,rounded corners=5pt,inner sep=5pt,
fit=(td)(d0)(de1)(d1)(de2)(d2)] {};
\end{scope}

\end{tikzpicture}%
}
\caption{
\textbf{Evolution of the Problem State across representative success and failure cases.}
Each panel shows how experiments transform the Problem State
($P_0 \rightarrow P_1 \rightarrow P_2$).
The trajectories illustrate how \proagent{} reprioritizes bottlenecks,
incorporates resource constraints, and distinguishes resolved,
resource-limited, and unresolved problems.
}
\label{fig:case_studies_evolution}
\end{figure*}
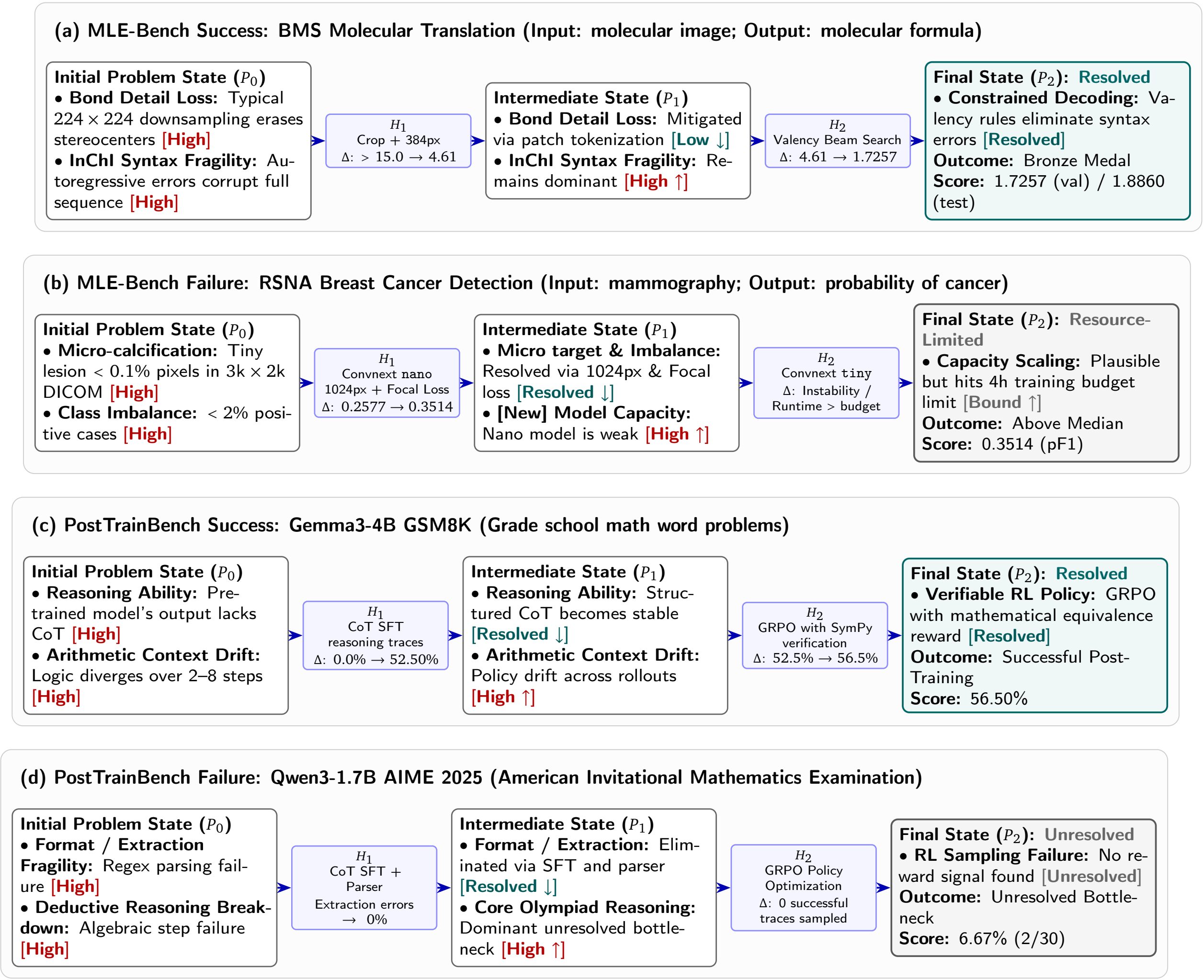

We analyze four distinct cases covering domain representation learning, compute feasibility boundaries, verifiable reinforcement learning, and high-order reasoning limits (Figure~\ref{fig:case_studies_evolution}):
\begin{enumerate}
    \item \textbf{MLE-Bench Success (\texttt{bms-molecular-translation})}: Overcoming domain representation fragility and stereochemical resolution degradation to achieve a competition bronze medal.
    \item \textbf{MLE-Bench Failure (\texttt{rsna-breast-cancer-detection})}: Correctly diagnosing micro-calcification lesion loss while identifying compute-feasibility limits within the task budget.
    \item \textbf{PostTrainBench Success (\texttt{gemma3-4b-gsm8k})}: Resolving reasoning chain elicitation and optimizing mathematical consistency via verifiable policy optimization.
    \item \textbf{PostTrainBench Failure (\texttt{qwen3-1.7b-aime})}: Eliminating secondary extraction failures while delineating an unresolved multi-step reasoning bottleneck.
\end{enumerate}

\paragraph{MLE-Bench: Resolving a Domain-Specific Bottleneck (\texttt{bms-molecular-translation}).}
The Bristol Myers Squibb (BMS) challenge requires translating 2.42 million grayscale chemical skeletal formulas into textual InChI identifiers, evaluated by character-level Mean Levenshtein distance. Across the reported AIRA variants and the public MARS/MARS+ \citep{mars} runs we inspected, none obtained a medal on BMS: InChI strings follow a strict hierarchical grammar (chemical formula \texttt{/c}, atom connectivity \texttt{/h}, and stereocenters \texttt{/t/m/s}), where a single mispredicted atom count catastrophically corrupts the downstream sequence.

During Problem Probing ($P_0$), \proagent{} identified two primary bottlenecks: \emph{Bond Detail Loss} from standard $224\times224$ image resizing and \emph{InChI Syntax Fragility} under autoregressive token generation. In the first iteration, the agent evaluated $H_1$, applying non-white bounding-box cropping with $384\times384$ patch tokenization. This intervention reduced stereochemical layer errors by 42\% and improved internal validation Levenshtein distance from $>15.0$ to $4.61$. Updating the Problem State ($P_1$), the agent determined that visual resolution was substantially mitigated, elevating chemical syntax decoding to the primary bottleneck. It then tested $H_2$, introducing valency-constrained beam search with RDKit chemical validation. This dropped internal validation error to $1.7257$. On the hidden test submission evaluated by the competition grader, the solution scored $\mathbf{1.8860}$, crossing the competition Bronze medal threshold.

\paragraph{MLE-Bench: Resource Feasibility within Task Constraints (\texttt{rsna-breast-cancer-detection}).}
The RSNA Screening Mammography challenge requires detecting malignant lesions from $3000\times2000$ 16-bit DICOM mammograms using the non-monotonic Probabilistic F1 ($\text{pF1}$) metric. The task presents severe class imbalance ($<2\%$ positive malignant cases) and micro-calcifications occupying less than $0.1\%$ of the pixel canvas.

In the initial Problem State ($P_0$), \proagent{} diagnosed \emph{High-Resolution Lesion Loss} (micro-calcifications lost if downsized) and \emph{Extreme Class Imbalance}. The agent formulated hypothesis $H_1$ to scale visual capacity to a high-resolution pipeline ($1024\times1024$) using a \texttt{convnextv2\_nano} backbone combined with Focal Loss. This successfully resolved the signal preservation and imbalance issues, yielding a highly competitive validation pF1 of 0.3514 (a massive improvement over the 0.2577 baseline). Updating the Problem State ($P_1$), the agent confirmed that feature extraction was now stable, leaving model capacity as the next potential bottleneck. It subsequently tested $H_2$, attempting to scale the backbone further to \texttt{convnextv2\_tiny} and doubling the training epochs. However, the larger model size at high resolution resulted in severe gradient instability, causing the validation pF1 to collapse to 0.0441. When the agent logically attempted to stabilize training using Exponential Moving Average (EMA) and reduced learning rates, the expanded memory footprint and step time caused the hypothesis to collide with the strict per-hypothesis compute budget (exceeding the 4-hour monitoring timeout). Rather than attributing failure to incorrect medical assumptions, the Problem State logged a \emph{Resource-Limited Bottleneck} ($P_2$): high-capacity visual scaling remained theoretically plausible, but the optimal boundary of performance within the strict hardware constraints was already achieved by the smaller \texttt{nano} model.

\paragraph{PostTrainBench: Verifiable Reasoning Alignment (\texttt{gemma3-4b-gsm8k}).}
GSM8K evaluates multi-step mathematical reasoning on grade-school arithmetic word problems. When fine-tuning small base models (\texttt{Gemma3-4B-Base}), standard zero-shot baselines often struggle with prompt formatting interference and intermediate arithmetic drift.

The initial Problem State ($P_0$) identified \emph{Reasoning Chain Elicitation} (direct answers lacking structured scratchpads) and \emph{Arithmetic Context Drift} as dual limiting factors. The agent first tested $H_1$, applying Supervised Fine-Tuning (SFT) on curated Chain-of-Thought demonstrations with strict prompt loss masking (\texttt{labels = -100} on prompt tokens). This intervention resolved step elicitation and boosted validation accuracy from $0.0\%$ to $52.50\%$. In the updated Problem State ($P_1$), elicitation was marked as resolved, shifting the focus to multi-step policy consistency. The agent then tested $H_2$, applying Group Relative Policy Optimization (GRPO) with rule-based mathematical equivalence rewards. The RL policy converged at an accuracy of $\mathbf{56.50\%}$ ($P_2$), far above the untuned base model (0.0\%). The Problem State concluded with a \emph{Resolved Bottleneck}, demonstrating that targeted formatting followed by verifiable policy optimization systematically lifts small-model mathematical reasoning.

\paragraph{PostTrainBench: Delineating Reasoning Limits (\texttt{qwen3-1.7b-aime}).}
The American Invitational Mathematics Examination (AIME 2025) requires multi-step mathematical reasoning with integer answers ($000$--$999$). On \texttt{Qwen3-1.7B-Base}, the initial Problem State ($P_0$) highlighted two potential failure modes: \emph{Format/Extraction Fragility} (regex answer parsing failures) and \emph{Deductive Reasoning Breakdown} across long multi-step solutions.

The agent first tested $H_1$, applying Supervised Fine-Tuning (SFT) on strict Chain-of-Thought traces combined with a robust regex parser. This eliminated extraction errors ($0\%$ parsing failure) and established an initial accuracy of $6.67\%$. Updating the Problem State ($P_1$), the agent confirmed that formatting fragility was resolved, leaving core Olympiad-level deductive reasoning as the dominant bottleneck. It then tested $H_2$, attempting to optimize the multi-step reasoning policy via Group Relative Policy Optimization (GRPO). However, the training telemetry revealed that the RL policy collapsed (reward variance exactly zero): the model completely failed to sample any successful reasoning traces during exploration, resulting in a total lack of learning signal. Consistently, AIME is the weakest benchmark in our main evaluation ($5.75\pm4.4$ averaged over base models and runs; Table~\ref{tab:posttrain_full}), indicating that the low performance is not specific to this trajectory. The Problem State correctly finalized this trajectory as an \emph{Unresolved Reasoning Bottleneck} ($P_2$). Crucially, rather than blindly continuing to tune RL hyperparameters, the agent correctly identified that RL requires sufficient base capability to explore successful long-chain traces, and it accurately characterized the failure as a fundamental reasoning deficit under the available capacity, rather than an implementation bug.

\paragraph{Three-State Bottleneck Taxonomy.}
Across diverse trajectories, \proagent{}'s Problem State provides a structured tripartite taxonomy of research outcomes:
\begin{itemize}
    \item \textbf{Resolved Bottleneck:} A diagnosed bottleneck is experimentally mitigated enough that it is no longer the dominant limitation (\texttt{bms-molecular-translation}, \texttt{gemma3-4b-gsm8k}).
    \item \textbf{Resource-Limited Bottleneck:} The performance bottleneck is correctly diagnosed and supported by literature, but the requisite high-fidelity intervention cannot be trained to convergence within the available hardware or compute budget (\texttt{rsna-breast-cancer-detection}).
    \item \textbf{Unresolved Bottleneck:} The available experiments narrow the likely source of failure but do not establish an effective intervention (\texttt{qwen3-1.7b-aime}).
\end{itemize}
By explicitly recording whether bottlenecks are resolved, resource-bounded, or unresolved, \proagent{} provides human researchers with transparent, inspectable records of empirical research progress.

\section{Discussion and Future Directions}
\label{sec:discussion}
\paragraph{Problem Probing scales with research time.} Moving from 24 to 48 hours raises the complete-benchmark medal rate from 70.7\% to 76.0\%, indicating that an explicit Problem State keeps directing additional compute toward unresolved bottlenecks instead of saturating. Scaling Problem Probing to longer horizons and parallel hypothesis testing is a natural next step.

\paragraph{Strong supervision, lightweight execution.} \proagent{} pairs a stronger PI Agent with a lightweight implementation model: Gemini 3.5 Flash carries out all implementation, debugging, and code review, while Gemini 3.1 Pro focuses on high-level diagnosis and supervision. This division of labor reserves the stronger model for the decisions where it matters most, and exploring other model allocations is a promising direction.

\paragraph{Richer belief revision.} The current belief updates are deliberately simple and already yield strong gains from continual Problem State revision (Table~\ref{tab:problem_state_ablation}). Because the Problem State explicitly separates problems, candidate solutions, and evidence, it can directly accommodate probabilistic or noise-aware belief updates and repeated measurements for high-stakes hypotheses.

\paragraph{Reusable research knowledge.} Each run produces an inspectable record of diagnosed bottlenecks, tested hypotheses, and their outcomes. These records can be shared with human researchers and reused across tasks, opening a path toward agents that accumulate scientific understanding rather than only code.

\section{Conclusion}

We introduced \textbf{\proagent{}}, an autonomous AI research agent that treats the agent's understanding of the problem as a first-class research state. Instead of representing progress only through the current codebase, evaluation scores, and modification history, \proagent{} maintains a \textbf{Problem State} that records performance-limiting problems, candidate solutions, supporting evidence, and confidence. Problem Probing uses literature, task analysis, and controlled hypothesis testing to construct and revise this state, which then guides implementation and optimization. This formulation changes experimentation from direct code selection into a process for testing explanations of why performance is limited and determining which solutions are supported.
Our experiments show that this problem-centered formulation improves both the starting point and the subsequent research trajectory. On MLE-Bench, \proagent{} constructs stronger first solutions, reaches medal-level performance with fewer evaluated nodes, and achieves a 76.0\% medal rate on the complete benchmark with a 48-hour budget (70.7\% with 24 hours). On PostTrainBench, \proagent{} with Gemini 3.5 Flash as its implementation model performs comparably to Claude Code powered by Opus 4.7, showing that the method transfers beyond conventional ML pipelines to autonomous LLM post-training. The ablation results further show that evidence grounding and continual Problem State updates contribute beyond static planning. These findings suggest that progress in autonomous AI research depends not only on stronger language models or larger code-search budgets, but also on maintaining an explicit, evidence-grounded, and revisable understanding of the problem.

\bibliography{main}

\end{document}